\documentclass[conference]{IEEEtran}
\usepackage{balance}
\usepackage{multicol}
\usepackage{amsmath}
\usepackage{mathrsfs}
\usepackage{amsfonts}
\usepackage{amssymb}
\usepackage{amsthm}

\newtheorem{theorem}{Theorem}

\ifCLASSINFOpdf
\else
\fi
\begin{document}
\title{{\Large \textbf{On the clustering aspect of nonnegative matrix factorization}}}

\author{\IEEEauthorblockN{Andri Mirzal}
\IEEEauthorblockA{Grad.~School of Information Science and Technology,\\
Hokkaido University, Kita 14 Nishi 9, Kita-Ku,\\
Sapporo, Japan\\
andri@complex.eng.hokudai.ac.jp}
\and
\IEEEauthorblockN{Masashi Furukawa}
\IEEEauthorblockA{Grad.~School of Information Science and Technology,\\
Hokkaido University, Kita 14 Nishi 9, Kita-Ku,\\
Sapporo, Japan\\
mack@complex.eng.hokudai.ac.jp}
}

\maketitle

\begin{abstract}
This paper provides a theoretical explanation on the clustering aspect of nonnegative matrix factorization (NMF). We prove that even without imposing orthogonality nor sparsity constraint on the basis and/or coef\mbox{}ficient matrix, NMF still can give clustering results, thus providing a theoretical support for many works, e.g., Xu et al.~\cite{Xu} and Kim et al.~\cite{Kim}, that show the superiority of the standard NMF as a clustering method.\\
\end{abstract}


\IEEEpeerreviewmaketitle

\section{Introduction} \label{introduction}
NMF is a matrix approximation technique that factorizes a nonnegative matrix into a pair of other nonnegative matrices of much lower rank:
\begin{equation}
\mathbf{A} \approx \mathbf{B}\mathbf{C},
\label{eq1}
\end{equation}
where $\mathbf{A}\in\mathbb{R}_{+}^{M\times N}=\left[\mathbf{a}_1,\ldots,\mathbf{a}_N\right]$ denotes the feature-by-item data matrix, $\mathbf{B}\in\mathbb{R}_{+}^{M\times K}=\left[\mathbf{b}_1,\ldots,\mathbf{b}_K\right]$ denotes the basis matrix, $\mathbf{C}\in\mathbb{R}_{+}^{K\times N}=\left[\mathbf{c}_1,\ldots,\mathbf{c}_N\right]$ denotes the coef\mbox{}ficient matrix, and $K$ denotes the number of factors which usually chosen so that $K\ll\min(M,N)$. There are also other variants of NMF like semi-NMF, convex NMF, and symmetric NMF. Detailed discussions can be found in, e.g., \cite{Li} and \cite{Ding3}.

The nonnegativity constraints and the reduced dimensionality define the uniqueness and power of NMF. The nonnegativity constraints allow only nonsubstractive linear combinations of the basis vectors $\mathbf{b}_k$ to construct the data vectors $\mathbf{a}_n$, thus providing the parts-based interpretations as shown in \cite{Lee, SZLi, Hoyer}. And the reduced dimensionality provides NMF with the clustering aspect and data compression capabilities. 

The most important NMF's application is in the data clustering, as some works have shown that it is a superior method compared to the standard clustering methods like spectral methods and $K$-means algorithm. In particular, Xu et al.~\cite{Xu} showed that NMF outperforms standard spectral methods in finding the document clustering in two text corpora, TDT2 and Reuters. And Kim et al.~\cite{Kim} showed that NMF and sparse NMF are much more superior methods compared to the $K$-means algorithm in both a synthetic dataset (which is well separated) and a real dataset (TDT2).

If sparsity constraints are imposed to columns of $\mathbf{C}$, the clustering aspect of NMF is intuitive since in the extreme case where there is only one nonzero entry per column, NMF will be equivalent to the $K$-means algorithm employed to the data vectors $\mathbf{a}_n$ \cite{Ding}, and the sparsity constraints can be thought as the relaxation to the strict orthogonality constraints on rows of $\mathbf{C}$ (an equivalent explanation can also be stated for imposing sparsity on rows of $\mathbf{B}$).

However, as reported by Xu et al.~\cite{Xu} and Kim et al.~\cite{Kim}, even without imposing sparsity constraints, NMF still can give very promising clustering results. But the authors didn't give any theoretical analysis on why the standard NMF---NMF without sparsity nor orthogonality constraint---can give such good results. So far the best explanation for this remarkable fact is only qualitative: the standard NMF produces non-orthogonal latent semantic directions (the basis vectors) that are more likely to correspond to each of the clusters than those produced by the spectral methods, thus the clustering induced from the latent semantic directions of the standard NMF are better than clustering by the spectral methods \cite{Xu}. Therefore, this work attempts to provide a theoretical support for the clustering aspect of the standard NMF.

\section{Clustering aspect of NMF} \label{clusteringnmf}
To compute $\mathbf{B}$ and $\mathbf{C}$, usually eq.~\ref{eq1} is rewritten into a minimization problem in the Frobenius norm criterion.
\begin{equation}
\min_{\mathbf{B},\mathbf{C}}J\left(\mathbf{B},\mathbf{C}\right)=\frac{1}{2}\|\mathbf{A}-\mathbf{B}\mathbf{C}\|_{F}^{2}\;\,\mathrm{s.t.}\;\, \mathbf{B}\ge\mathbf{0},\mathbf{C}\ge\mathbf{0}.
\label{eq2}
\end{equation}
In addition to the usual Frobenius norm criterion, the family of Bregman divergences---which Frobenius norm and Kullback-Leibler divergence are part of it---can also be used as the af\mbox{}finity measures. Detailed discussion on the Bregman divergences for NMF can be found in \cite{Dhillon}.

Sometimes it is more practical and intuitive to decompose $J\left(\mathbf{B},\mathbf{C}\right)$ into a series of smaller objectives.
\begin{align}
&\min_{\mathbf{B},\mathbf{C}}J\left(\mathbf{B},\mathbf{C}\right)\equiv \left(\min_{\mathbf{B},\mathbf{c}_1}J_1(\mathbf{B},\mathbf{c}_1),\ldots,\min_{\mathbf{B},\mathbf{c}_N}J_N(\mathbf{B},\mathbf{c}_N)\right), \label{eq000}\\
&\text{where} \nonumber \\
&\min_{\mathbf{B},\mathbf{c}_n}J_n\left(\mathbf{B},\mathbf{c}_n\right)=\frac{1}{2}\|\mathbf{a}_n-\mathbf{B}\mathbf{c}_n\|_2^2,\;\,n\in[1,N]. \label{eq010}
\end{align}

Minimizing $J_n$ is known to be the nonnegative least square (NLS) problem, and some fast NMF algorithms are developed based on solving the NLS subproblems, e.g., alternating NLS with block principal pivoting algorithm \cite{Kim2}, active set method \cite{HKim}, and projected quasi-Newton algorithm \cite{DKim}. Decomposing NMF problem into NLS subproblems also transforms the non-convex optimization in eq.~\ref{eq000} to the convex optimization subproblems in eq.~\ref{eq010}. Even though eq.~\ref{eq010} is not strictly convex, for two-block case, any limit point of the sequence \{$\mathbf{B}^t$,$\mathbf{C}^t$\}, where $t$ is the updating step, is a stationary point \cite{Grippo}.

The objective in eq.~\ref{eq010} aims to simultaneously find the suitable basis vectors such that the latent factors are revealed, and the coef\mbox{}ficient vector $\mathbf{c}_n$ such that a linear combination of the basis vectors ($\mathbf{B}\mathbf{c}_n$) is close to $\mathbf{a}_n$. In clustering term this can be rephrased as: to simultaneously find the cluster centers and the cluster assignments.

To investigate the clustering aspect of NMF, four possibilities of NMF settings are discussed: (1) imposing orthogonality constraints on both rows of $\mathbf{C}$ and columns of $\mathbf{B}$, (2) imposing orthogonality constraints on rows of $\mathbf{C}$, (3) imposing orthogonality constraints on columns of $\mathbf{B}$, and (4) no orthogonality constraint is imposed. The last case is the standard NMF which its clustering aspect is the focus of this paper as many works reported that it is a very ef\mbox{}fective clustering method.

\subsection{Orthogonality constraints on both $\mathbf{B}$ and $\mathbf{C}$} \label{BC}
The following theorems proves that imposing column-orthogonality constraints on $\mathbf{B}$ and row-orthogonality constraints on $\mathbf{C}$ lead to the simultaneous clustering of similar items and related features.
\begin{theorem} \label{theorem1}
Minimizing the following objective
\begin{eqnarray}
&&\min_{\mathbf{B},\mathbf{C}}J_a\left(\mathbf{B},\mathbf{C}\right)=\frac{1}{2}\|\mathbf{A}-\mathbf{B}\mathbf{C}\|_{F}^{2} \label{eqa}\\
&&\mathrm{s.t.}\;\,\mathbf{B}\ge\mathbf{0},\mathbf{C}\ge\mathbf{0},\mathbf{B}^T\mathbf{B}=\mathbf{I}, \mathbf{C}\mathbf{C}^T=\mathbf{I} \nonumber
\label{eqb}
\end{eqnarray}
is equivalent to applying ratio association to $\mathcal{G}(\mathbf{A}^T\mathbf{A})$ and $\mathcal{G}(\mathbf{A}\mathbf{A}^T)$, where $\mathbf{A}^T\mathbf{A}$ and $\mathbf{A}\mathbf{A}^T$ are the item af\mbox{}finity matrix and the feature af\mbox{}finity matrix respectively, thus leads to simultaneous clustering of similar items and related features.
\end{theorem}
\begin{IEEEproof}
\begin{align}
\|\mathbf{A}-\mathbf{B}\mathbf{C}\|_{F}^{2}=\;&\mathrm{tr}\left(\left(\mathbf{A}-\mathbf{B}\mathbf{C}\right)^T\left(\mathbf{A}-\mathbf{B}\mathbf{C}\right)\right) \nonumber \\
=\;&\mathrm{tr}\left(\mathbf{A}^T\mathbf{A}-2\mathbf{C}^T\mathbf{B}^T\mathbf{A}+\mathbf{I}\right). \label{eqc}
\end{align}
The Lagrangian function:
\begin{align}
L_a\left(\mathbf{B},\mathbf{C}\right)=\;&J_a\left(\mathbf{B},\mathbf{C}\right)-\mathrm{tr}\left(\mathbf{\Gamma}_{\mathbf{B}}\mathbf{B}^T\right)-\mathrm{tr}\left(\mathbf{\Gamma}_{\mathbf{C}}\mathbf{C}\right) + \nonumber \\
&\mathrm{tr}\left(\mathbf{\Lambda}_{\mathbf{B}}\left(\mathbf{B}^T\mathbf{B}-\mathbf{I}\right)\right)+\mathrm{tr}\left(\mathbf{\Lambda}_{\mathbf{C}}\left(\mathbf{C}\mathbf{C}^T-\mathbf{I}\right)\right), \label{eqaa}
\end{align}
where $\mathbf{\Gamma}_{\mathbf{B}}\in\mathbb{R}_{+}^{M\times K}$, $\mathbf{\Gamma}_{\mathbf{C}}\in\mathbb{R}_{+}^{N\times K}$, $\mathbf{\Lambda}_{\mathbf{B}}\in\mathbb{R}_{+}^{K\times K}$, and $\mathbf{\Lambda}_{\mathbf{C}}\in\mathbb{R}_{+}^{K\times K}$ are the Lagrange multipliers. By the Karush-Kuhn-Tucker (KKT) optimality conditions we get:
\begin{align}
\nabla_{\mathbf{B}}L_a=\;&\mathbf{B}-\mathbf{A}\mathbf{C}^T-\mathbf{\Gamma}_{\mathbf{B}}+2\mathbf{B}\mathbf{\Lambda}_{\mathbf{B}}=\mathbf{0},\label{eqbb} \\
\nabla_{\mathbf{C}}L_a=\;&\mathbf{C}-\mathbf{B}^T\mathbf{A}-\mathbf{\Gamma}_{\mathbf{C}}^T+2\mathbf{\Lambda}_{\mathbf{C}}\mathbf{C}=\mathbf{0},\label{eqcc}
\end{align}
with complementary slackness:
\begin{equation}
\mathbf{\Gamma}_{\mathbf{B}}\otimes\mathbf{B}=\mathbf{0},\;\,\mathbf{\Gamma}_{\mathbf{C}}^T\otimes\mathbf{C}=\mathbf{0}, \label{eqdd}
\end{equation}
where $\otimes$ denotes component-wise multiplications. Assume $\mathbf{\Gamma}_{\mathbf{B}}=\mathbf{0}$, $\mathbf{\Lambda}_{\mathbf{B}}=\mathbf{0}$, $\mathbf{\Gamma}_{\mathbf{C}}=\mathbf{0}$, and $\mathbf{\Lambda}_{\mathbf{B}}=\mathbf{0}$ (at the stationary point these assumptions are reasonable since the complementary slackness conditions hold and the Lagrange multipliers can be assigned to zeros), we get:
\begin{align}
\mathbf{B}=\;&\mathbf{A}\mathbf{C}^T\;\text{and} \label{eqee}\\
\mathbf{C}=\;&\mathbf{B}^T\mathbf{A}. \label{eqff}
\end{align}
Substituting eq.~\ref{eqee} into eq.~\ref{eqc}, we get: 
\begin{equation}
\min_{\mathbf{C}}J_a\left(\mathbf{C}\right) \equiv \max_{\mathbf{C}}\;\mathrm{tr}\left(\mathbf{C}\mathbf{A}^T\mathbf{A}\mathbf{C}^T\right).
\label{eqf}
\end{equation}
Similarly, substituting eq.~\ref{eqff} into eq.~\ref{eqc}, we get:
\begin{equation}
\min_{\mathbf{B}}J_a\left(\mathbf{B}\right) \equiv \max_{\mathbf{B}}\;\mathrm{tr}\left(\mathbf{B}^T\mathbf{A}\mathbf{A}^T\mathbf{B}\right).
\label{eqe}
\end{equation}
Therefore, minimizing $J_a$ is equivalent to simultaneously optimizing:
\begin{align}
&\max_{\mathbf{C}}\;\mathrm{tr}\left(\mathbf{C}\mathbf{A}^T\mathbf{A}\mathbf{C}^T\right)\;\,\mathrm{s.t.}\;\,\mathbf{C}\mathbf{C}^T=\mathbf{I},\;\,\mathrm{and} \label{eqg}\\
&\max_{\mathbf{B}}\;\mathrm{tr}\left(\mathbf{B}^T\mathbf{A}\mathbf{A}^T\mathbf{B}\right)\;\,\mathrm{s.t.}\;\,\mathbf{B}^T\mathbf{B}=\mathbf{I}.\label{eq2s4}
\end{align}
Eq.~~\ref{eqg} and eq.\ref{eq2s4} are the ratio association objectives (see \cite{Dhillon1} for details on various graph cuts objectives) applied to $\mathcal{G}(\mathbf{A}^T\mathbf{A})$ and $\mathcal{G}(\mathbf{A}\mathbf{A})^T$ respectively. Thus minimizing $J_a$ leads to the simultaneous clustering of similar items and related features.
\end{IEEEproof}

\subsection{Orthogonality constraints on $\mathbf{C}$} \label{C}
When the orthogonality constraints are imposed only on rows of $\mathbf{C}$, it is no longer clear whether columns of $\mathbf{B}$ will lead to the feature clustering. The following theorem shows that without imposing the orthogonality constraints on $\mathbf{b}_k$, the resulting $\mathbf{B}$ can still lead to the feature clustering.
\begin{theorem} \label{theorem2}
Minimizing the following objective
\begin{align}
&\min_{\mathbf{B},\mathbf{C}}J_{b}\left(\mathbf{B},\mathbf{C}\right)=\frac{1}{2}\|\mathbf{A}-\mathbf{B}\mathbf{C}\|_{F}^{2} \label{eq4}\\
&\mathrm{s.t.}\;\,\mathbf{B}\ge\mathbf{0},\mathbf{C}\ge\mathbf{0},\mathbf{C}\mathbf{C}^T=\mathbf{I} \nonumber 
\end{align}
is equivalent to applying ratio association to $\mathcal{G}(\mathbf{A}^T\mathbf{A})$, and also leads to the feature clustering indicator matrix $\mathbf{B}$ which is approximately column-orthogonal.
\end{theorem}
\begin{IEEEproof}
\begin{align}
\|\mathbf{A}-\mathbf{B}\mathbf{C}\|_{F}^{2}=\;&\mathrm{tr}\left(\left(\mathbf{A}-\mathbf{B}\mathbf{C}\right)^T\left(\mathbf{A}-\mathbf{B}\mathbf{C}\right)\right) \nonumber \\
=\;&\mathrm{tr}\left(\mathbf{A}^T\mathbf{A}-2\mathbf{B}^T\mathbf{A}\mathbf{C}^T+\mathbf{C}^T\mathbf{B}^T\mathbf{B}\mathbf{C}\right). \label{eq6a}
\end{align}
The Lagrangian function:
\begin{align}
L_b\left(\mathbf{B},\mathbf{C}\right)=\;&J_b\left(\mathbf{B},\mathbf{C}\right)-\mathrm{tr}\left(\mathbf{\Gamma}_{\mathbf{B}}\mathbf{B}^T\right)-\mathrm{tr}\left(\mathbf{\Gamma}_{\mathbf{C}}\mathbf{C}\right) + \nonumber \\
&\mathrm{tr}\left(\mathbf{\Lambda}_{\mathbf{C}}\left(\mathbf{C}\mathbf{C}^T-\mathbf{I}\right)\right). \label{eq7a}
\end{align}
By applying the KKT conditions, we get:
\begin{align}
\mathbf{B}=\;&\mathbf{A}\mathbf{C}^T\;\text{and} \label{eq8a}\\
\mathbf{C}=\;&\mathbf{B}^T\mathbf{A}. \label{eq9a}
\end{align}
By substituting eq.~\ref{eq8a} and eq.~\ref{eq9a} into eq.~\ref{eq6a}, minimizing $J_b$ is equivalent to simultaneously optimizing:
\begin{align}
&\max_{\mathbf{C}}\;\mathrm{tr}\left(\mathbf{C}\mathbf{A}^T\mathbf{A}\mathbf{C}^T\right)\;\,\text{s.t.}\;\,\mathbf{C}\mathbf{C}^T=\mathbf{I}, \label{eq10a}\\
&\max_{\mathbf{B}}\;\mathrm{tr}\left(\mathbf{B}^T\mathbf{A}\mathbf{A}^T\mathbf{B}\right), \label{eq11a}\;\,\text{and} \\
&\min_{\mathbf{B}}\;\mathrm{tr}\left(\mathbf{A}^T\mathbf{B}\mathbf{B}^T\mathbf{B}\mathbf{B}^T\mathbf{A}\right)\equiv\min_{\mathbf{B}}\;\mathrm{tr}\left(\mathbf{B}^T\mathbf{B}\mathbf{B}^T\mathbf{B}\right).
\label{eq12a}
\end{align}
Note that the step in eq.~\ref{eq12a} is justifiable since $\mathbf{A}$ is a constant matrix. By using the fact $\mathrm{tr}(\mathbf{X}^T\mathbf{X})=\|\mathbf{X}\|_F^2$, eq.~\ref{eq12a} can be rewritten as:
\begin{equation}
\min_{\mathbf{B}}\;\left\|\mathbf{B}^T\mathbf{B}\right\|_F^2=\min_{\mathbf{B}}\;\Big(\sum_{i}\left(\mathbf{b}_{i}^T\mathbf{b}_{i}\right)^2 + \sum_{i\ne j}\left(\mathbf{b}_{i}^T\mathbf{b}_{j}\right)^2 \Big). \label{eq13abc}
\end{equation}

The objective in eq.~\ref{eq10a} is equivalent to eq.~\ref{eqg} and eventually leads to the clustering of similar items. So the remaining problem is how to prove that optimizing eq.~\ref{eq11a} and \ref{eq13abc} simultaneously will lead to the feature clustering indicator matrix $\mathbf{B}$ which is approximately column-orthogonal.

Eq.~\ref{eq11a} resembles eq.~\ref{eqe}, but without orthogonality nor upper bound constraint, so one can easily optimizing eq.~\ref{eq11a} by setting $\mathbf{B}$ to an infinity matrix. However, this violates eq.~\ref{eq13abc} which favors small $\mathbf{B}$. Conversely, one can optimizing eq.~\ref{eq13abc} by setting $\mathbf{B}$ to a null matrix, but again this violates eq.~\ref{eq11a}. Therefore, these two objectives create implicit lower and upper bound constraints on $\mathbf{B}$, and eq.~\ref{eq11a} and eq.~\ref{eq13abc} can be rewritten into:
\begin{align}
&\max_{\mathbf{B}}\;\mathrm{tr}\left(\mathbf{B}^T\mathbf{\hat{A}}\mathbf{B}\right),\;\,\text{and}  \label{eq13aa} \\
&\min_{\mathbf{B}}\;\Big( \underbrace{\sum_{i}\left(\mathbf{b}_{i}^T\mathbf{b}_{i}\right)^2}_{j_{b1}} + \underbrace{\sum_{i\ne j}\left(\mathbf{b}_{i}^T\mathbf{b}_{j}\right)^2}_{j_{b2}} \Big) \label{eq13ab} \\
&\text{s.t.}\;\,\mathbf{0}\le\mathbf{B}\le\mathbf{\Upsilon}_{\mathbf{B}}, \nonumber
\end{align}
where $\mathbf{\hat{A}}$ denotes the feature affinity matrix and $\mathbf{\Upsilon}_{\mathbf{B}}$ denotes the upperbound constraints on $\mathbf{B}$. Now we have box-constraint objectives which are known to behave well and are guaranteed to converge to the stationary point \cite{Calamai}. 

Even though the objectives are now transformed into box-constraint optimization problems, since there is no column-orthogonality constraint, maximizing eq.~\ref{eq13aa} can be easily done by setting each entry of $\mathbf{B}$ to the corresponding largest possible value (in graph term this means to only create one partition on $\mathcal{G}(\mathbf{\hat{A}})$). But this scenario results in the maximum value of eq.~\ref{eq13ab}, which violates the objective. Conversely, minimizing eq.~\ref{eq13ab} to the smallest possible value (minimizing $j_{b1}$ implies minimizing $j_{b2}$, but not vice versa) violates eq.~\ref{eq13aa}. 

Thus, the most reasonable scenario is: setting $j_{b2}$ as small as possible and balancing $j_{b1}$ with eq.~\ref{eq13aa}. This scenario is the relaxed ratio association applied to $\mathcal{G}(\mathbf{\hat{A}})$, and as long as vertices of $\mathcal{G}(\mathbf{\hat{A}})$ are clustered, simultaneous optimizing eq.~\ref{eq13aa} and eq.~\ref{eq13ab} leads to the clustering of related features. Moreover, as $j_{b2}$ is minimum, $\mathbf{B}$ is approximately column-orthogonal.
\end{IEEEproof}

\subsection{Orthogonality constraints on $\mathbf{B}$} \label{B}
\begin{theorem} \label{theorem3}
Minimizing the following objective
\begin{align}
&\min_{\mathbf{B},\mathbf{C}}J_{c}\left(\mathbf{B},\mathbf{C}\right)=\frac{1}{2}\|\mathbf{A}-\mathbf{B}\mathbf{C}\|_{F}^{2} \label{eq13}\\
&\mathrm{s.t.}\;\,\mathbf{B}\ge\mathbf{0},\mathbf{C}\ge\mathbf{0},\mathbf{B}^T\mathbf{B}=\mathbf{I} \nonumber 
\end{align}
is equivalent to applying ratio association to $\mathcal{G}(\mathbf{A}\mathbf{A}^T)$, and also leads to the item clustering indicator matrix $\mathbf{C}$ which is approximately row-orthogonal.
\end{theorem}
\begin{IEEEproof}
By following the proof of theorem \ref{theorem2}, minimizing $J_c$ is equivalent to simultaneously optimizing:
\begin{align}
&\max_{\mathbf{B}}\;\mathrm{tr}\left(\mathbf{B}^T\mathbf{A}\mathbf{A}^T\mathbf{B}\right)\;\,\text{s.t.}\;\,\mathbf{B}^T\mathbf{B}=\mathbf{I}, \label{eq14a}\\
&\max_{\mathbf{C}}\;\mathrm{tr}\left(2\mathbf{C}^T\mathbf{A}^T\mathbf{A}\mathbf{C}\right),\;\,\text{and} \label{eq14b}\\
&\min_{\mathbf{C}}\;\mathrm{tr}\left(\mathbf{C}^T\mathbf{C}\mathbf{A}^T\mathbf{A}\mathbf{C}^T\mathbf{C}\right)\equiv\min_{\mathbf{C}}\;\mathrm{tr}\left(\mathbf{C}\mathbf{C}^T\mathbf{C}\mathbf{C}^T\right).
\label{eq14c}
\end{align}

Eq.~\ref{eq14a} is equivalent to eq.~\ref{eq2s4} and leads to the clustering of related features. And optimizing eq.~\ref{eq14b} and Eq.~\ref{eq14c} simultaneously is equivalent to: 
\begin{align}
&\max_{\mathbf{C}}\;\mathrm{tr}\left(\mathbf{C}\mathbf{\tilde{A}}\mathbf{C}^T\right),\;\,\text{and}  \label{eq14aa} \\
&\min_{\mathbf{C}}\;\Big(\underbrace{\sum_{i}\left(\mathbf{\check{c}}_{i}\mathbf{\check{c}}_{i}^T\right)^2}_{j_{c1}} + \underbrace{\sum_{i\ne j}\left(\mathbf{\check{c}}_{i}\mathbf{\check{c}}_{j}^T\right)^2}_{j_{c2}} \Big) \label{eq14ab} \\
&\text{s.t.}\;\,\mathbf{0}\le\mathbf{C}\le\mathbf{\Upsilon}_{\mathbf{C}}, \nonumber
\end{align}
where $\mathbf{\tilde{A}}$ denotes the item affinity matrix, $\mathbf{\check{c}}_i$ denotes the $i$-th row of $\mathbf{C}$, and $\mathbf{\Upsilon}_{\mathbf{C}}$ denotes the upperbound constraints on $\mathbf{C}$. 

As in the proof of theorem \ref{theorem2}, the most reasonable scenario in simultaneously optimizing eq.~\ref{eq14aa} and eq.~\ref{eq14ab} is by setting $j_{c2}$ as small as possible and balancing $j_{c1}$ with eq.~\ref{eq14aa}. This leads to the clustering of similar items, and as $j_{c2}$ is minimum, $\mathbf{C}$ is approximately row-orthogonal.
\end{IEEEproof}

\subsection{No orthogonality constraint on both $\mathbf{B}$ and $\mathbf{C}$} \label{NO}
In this section we prove that applying the standard NMF to the feature-by-item data matrix eventually leads to the simultaneous feature and item clustering.
\begin{theorem} \label{theorem4}
Minimizing the following objective
\begin{align}
&\min_{\mathbf{B},\mathbf{C}}J_{d}\left(\mathbf{B},\mathbf{C}\right)=\frac{1}{2}\|\mathbf{A}-\mathbf{B}\mathbf{C}\|_{F}^{2} \label{eq15}\\
&\mathrm{s.t.}\;\,\mathbf{B}\ge\mathbf{0},\mathbf{C}\ge\mathbf{0}, \nonumber 
\end{align}
leads to the feature clustering indicator matrix $\mathbf{B}$ and the item clustering indicator matrix $\mathbf{C}$ which are approximately column- and row-orthogonal respectively.
\end{theorem}
\begin{IEEEproof}
By following the proof of theorem \ref{theorem2}, minimizing $J_d$ is equivalent to simultaneously optimizing:
\begin{align}
&\max_{\mathbf{B},\mathbf{C}}\mathrm{tr}\left(\mathbf{B}^T\mathbf{A}\mathbf{C}^T\right), \label{eq17a} \;\,\text{and} \\
&\min_{\mathbf{B},\mathbf{C}}\mathrm{tr}\left(\mathbf{B}^T\mathbf{B}\mathbf{C}\mathbf{C}^T\right). \label{eq17b}
\end{align}
By substituting $\mathbf{B}=\mathbf{A}\mathbf{C}^T$ and $\mathbf{C}=\mathbf{B}^T\mathbf{A}$ into the above equations, we get:
\begin{align}
&\max_{\mathbf{B}}\mathrm{tr}\left(\mathbf{B}^T\mathbf{\hat{A}}\mathbf{B}\right), \label{eq18a} \;\,\text{and} \\
&\min_{\mathbf{B}}\mathrm{tr}\left(\mathbf{B}^T\mathbf{B}\mathbf{B}^T\mathbf{A}\mathbf{A}^T\mathbf{B}\right)\equiv\min_{\mathbf{B}}\mathrm{tr}\left(\mathbf{B}^T\mathbf{B}\mathbf{B}^T\mathbf{B}\right) \label{eq18b}
\end{align}
for feature clustering, and:
\begin{align}
&\max_{\mathbf{C}}\mathrm{tr}\left(\mathbf{C}\mathbf{\tilde{A}}\mathbf{C}^T\right), \label{eq19a} \;\,\text{and} \\
&\min_{\mathbf{C}}\mathrm{tr}\left(\mathbf{C}\mathbf{A}^T\mathbf{A}\mathbf{C}^T\mathbf{C}\mathbf{C}^T\right)\equiv\min_{\mathbf{C}}\mathrm{tr}\left(\mathbf{C}\mathbf{C}^T\mathbf{C}\mathbf{C}^T\right) \label{eq19b}
\end{align}
for item clustering. Therefore, minimizing $J_d$ is equivalent to simultaneously optimizing:
\begin{align}
&\max_{\mathbf{B}}\mathrm{tr}\left(\mathbf{B}^T\mathbf{\hat{A}}\mathbf{B}\right),\label{eq20a}\\
&\min_{\mathbf{B}}\;\Big(\sum_{i}\left(\mathbf{b}_{i}^T\mathbf{b}_{i}\right)^2 + \sum_{i\ne j}\left(\mathbf{b}_{i}^T\mathbf{b}_{j}\right)^2\Big), \label{eq20b} \\
&\max_{\mathbf{C}}\mathrm{tr}\left(\mathbf{C}\mathbf{\tilde{A}}\mathbf{C}^T\right),\;\,\text{and}\label{eq20c} \\
&\min_{\mathbf{C}}\;\Big(\sum_{i}\left(\mathbf{\check{c}}_{i}\mathbf{\check{c}}_{i}^T\right)^2 + \sum_{i\ne j}\left(\mathbf{\check{c}}_{i}\mathbf{\check{c}}_{j}^T\right)^2\Big), \label{eq20d} \\
&\text{s.t.}\;\,\mathbf{0}\le\mathbf{B}\le\mathbf{\Upsilon}_{\mathbf{B}},\;\,\text{and}\;\,\mathbf{0}\le\mathbf{C}\le\mathbf{\Upsilon}_{\mathbf{C}}, \nonumber
\end{align}
which will lead to the feature clustering indicator matrix $\mathbf{B}$ and the item clustering indicator matrix $\mathbf{C}$ that are approximately column- and row-orthogonal respectively.
\end{IEEEproof}

\section{Unipartite and directed graph cases} \label{ud}
\balance
The af\mbox{}finity matrix $\mathbf{W}$ induced from a unipartite (undirected) graph is a symmetric matrix, which is a special case of the rectangular af\mbox{}finity matrix $\mathbf{A}$. Therefore, by following the discussion in section \ref{clusteringnmf}, it can be shown that the standard NMF applied to $\mathbf{W}$ leads to the clustering indicator matrix which is almost orthogonal.

The af\mbox{}finity matrix $\mathbf{V}$ induced from a directed graph is an asymmetric square matrix. Since columns and rows of $\mathbf{V}$ correspond to the same set of vertices with the same order, as the clustering problem is concerned, $\mathbf{V}$ can be replaced by $\mathbf{V}+\mathbf{V}^T$ which is a symmetric matrix. Then the standard NMF can be applied to this matrix to get the clustering indicator matrix which is almost orthogonal.

\section{Related works} \label{rw}
Ding et al.~\cite{Ding} provides the theoretical analysis on the equivalences between orthogonal NMF to $K$-means clustering for both rectangular data matrices and symmetric matrices. However as their proofs utilize the zero gradient conditions, the hidden assumptions (setting the Lagrange multipliers to zeros) are not revealed there. Actually it can be easily shown that their approach is the KKT conditions applied to the unconstrained version of eq.~\ref{eq2}. Thus there is no guarantee that minimizing eq.~\ref{eq2} by using the zero gradient conditions leads to the stationary point located on the nonnegative orthant as required by the objective.

Applying the standard NMF to the symmetric matrix leads to almost orthogonal matrix was previously proven by Ding et al.~\cite{Ding2}. But due to the used approach, the theorem cannot be extended to the rectangular matrices which so far are the usual form of the data (practical applications of NMF seemed exclusively for rectangular matrices). Therefore, their results cannot be used to explain the abundant experimental results that show the power of the standard NMF in clustering, latent factors identification, learning the parts of objects, and producing sparse matrices even without explicit sparsity constraint \cite{Lee}.

\section{Conclusion} \label{conc}
By using the strict KKT optimality conditions, we showed that even without explicitly imposing orthogonality nor sparsity constraint NMF produces approximately column-orthogonal basis matrix and row-orthogonal coefficient matrix which lead to the simultaneous feature and item clustering. This result, therefore, gives the theoretical explanation on some experimental results that show the power of the standard NMF as a clustering tool which are reported to be better than the spectral methods \cite{Xu} and $K$-means algorithm \cite{Kim}.


\begin{thebibliography}{1}

\bibitem{Xu} W.~Xu, X.~Liu and Y.~Gong, ``Document clustering based on non-negative matrix factorization,'' Proc.~ACM SIGIR, pp.~267-73, 2003.

\bibitem{Kim} J.~Kim and H.~Park, ``Sparse nonnegative matrix factorization for clustering,'' CSE Technical Reports, Georgia Institute of Technology, 2008.

\bibitem{Li} T.~Li and C.~Ding, ``The relationships among various nonnegative matrix factorization methods for clustering,'' Proc.~ACM 6th Int'l Conf.~on Data Mining, pp.~362-71, 2006.

\bibitem{Ding3} C.~Ding, T.~Li, and M.I.~Jordan, ``Convex and Semi-Nonnegative Matrix Factorizations,''  IEEE Transactions on Pattern Analysis and Machine Intelligence, pp.~45-55, 2010.

\bibitem{Lee} D.~Lee and H.~Seung, ``Learning the parts of objects by non-negative matrix factorization,'' Nature, 401(6755), pp.~788-91, 1999.

\bibitem{SZLi} S.Z.~Li, X.W.~Hou, H.J.~Zhang, and Q.S.~Cheng, ``Learning spatially localized, parts-based representation,'' Proc.~IEEE Comp.~Soc.~Conf.~on Computer Vision and Pattern Recognition, pp.~207-12, 2001.

\bibitem{Hoyer} P.O.~Hoyer, ``Non-negative Matrix Factorization with Sparseness Constraints,'' The Journal of Machine Learning Research, Vol.~5, pp.~1457-69, 2004.

\bibitem{Ding} C.~Ding, T.~Li, W.~Peng, and H.~Park, ``Orthogonal nonnegative matrix t-factorizations for clustering,'' Proc.~12th ACM SIGKDD Int'l Conf.~on Knowledge Discovery and Data Mining, pp.~126-35, 2006.

\bibitem{Dhillon} I.S.~Dhillon and S.~Sra, ``Generalized nonnegative matrix approximation with Bregman divergences,'' UTCS Technical Reports, The University of Texas at Austin, 2005.

\bibitem{Kim2} J.~Kim and H.~Park, ``Toward faster nonnegative matrix factorization: A new algorithm and comparisons,'' Proc.~8th IEEE International Conference on Data Mining, pp.~353-62, 2008.

\bibitem{HKim} H.~Kim and H.~Park, ``Nonnegative matrix factorization based on alternating nonnegativity constrained least squares and active set method,'' SIAM. J. Matrix Anal. \& Appl., Vol.~30(2), pp.~713-30, 2008.

\bibitem{DKim} D.~Kim, S.~Sra, and I.S.~Dhillon, ``Fast projection-based methods for the least squares nonnegative matrix approximation problem,'' Stat. Anal. Data Min., Vol.~1(1), pp.~38-51, 2008.

\bibitem{Grippo} L.~Grippo and M.~Sciandrone, ``On the convergence of the block nonlinear Gauss-Seidel method under convex constraints,'' Operation Research Letters, Vol.~26, pp.~127-36, 2000.

\bibitem{Dhillon1} I.~S.~Dhillon, Y.~Guan, and B.~Kulis, ``Weighted Graph Cuts without eigenvectors: A multilevel approach,'' IEEE Transactions on Pattern Analysis and Machine Intelligence, Vol.~29, No.~11, pp.~1944-57, 2007.


\bibitem{Calamai} P.H.~Calamai and J.J.~More, ``Projected gradient methods for linearly constrained problems,'' Mathematical Programming, Vol.~39, pp.~93-116, 1987.


\bibitem{Ding2} C.~Ding, X.~He, and H.D.~Simon, ``On the equivalence of nonnegative matrix factorization and spectral clustering,'' Proc.~SIAM Data Mining Conference, pp.~606-10, 2005.

\end{thebibliography}
\end{document}